\def\year{2021}\relax
\documentclass[letterpaper]{article} % DO NOT CHANGE THIS
\usepackage{aaai21}  % DO NOT CHANGE THIS
\usepackage{times}  % DO NOT CHANGE THIS
\usepackage{helvet} % DO NOT CHANGE THIS
\usepackage{courier}  % DO NOT CHANGE THIS
\usepackage[hyphens]{url}  % DO NOT CHANGE THIS
\usepackage{graphicx} % DO NOT CHANGE THIS
\usepackage{natbib}  % DO NOT CHANGE THIS AND DO NOT ADD ANY OPTIONS TO IT
\usepackage{caption} % DO NOT CHANGE THIS AND DO NOT ADD ANY OPTIONS TO IT
\usepackage{graphicx}
\usepackage{algorithm}  
\usepackage{algpseudocode}  
\usepackage{amsmath}
\usepackage{booktabs}
\usepackage{amsfonts}
\usepackage{subfigure}
\usepackage{makecell}

\title{PDLight: A Deep Reinforcement Learning Traffic Light Control Algorithm with Pressure and Dynamic Light Duration }
\author {
	Chenguang Zhao,\textsuperscript{\rm 1}
	Xiaorong Hu,\textsuperscript{\rm 1}
	Gang Wang \textsuperscript{\rm 1} \\
}
\affiliations {
	\textsuperscript{\rm 1} Department of Electronic and Information Engineering, Beihang University \\
	\{zchenguang,hxiaorong,gwang\}@buaa.edu.cn
}

\begin{document}	
\maketitle

\begin{abstract}
\noindent Existing ineffective and inflexible traffic light control at urban intersections can often lead to congestion in traffic flows and cause numerous problems, such as long delay and waste of energy. How to find the optimal signal timing strategy is a significant challenge in urban traffic management. In this paper, we propose PDlight, a deep reinforcement learning (DRL) traffic light control algorithm with a novel reward as PRCOL (Pressure with Remaining Capacity of Outgoing Lane). Serving as an improvement over the pressure used in traffic control algorithms, PRCOL considers not only the number of vehicles on the incoming lane but also the remaining capacity of the outgoing lane. Simulation results using both synthetic and real-world data-sets show that the proposed PDlight yields lower average travel time compared with several state-of-the-art algorithms, PressLight and Colight, under both fixed and dynamic green light duration.

\end{abstract}

\section{Introduction}
Traffic congestion is one of the major problems in today’s metropolitan transportation networks. Effective traffic signal control plays a significant role in alleviating traffic congestion, reducing the average travel time, improving throughput of a traffic network. However, the optimization of traffic lights scheduling at urban road intersections is a challenging task, especially how to intelligently adapt to real-time traffic demands.

Traditional traffic signal control systems usually use fixed phase and time schedule, which can only adjust the traffic flow at the intersection in a known traffic pattern. With increasing availability of large volumes of sensors, it is now potential to make significant improvements in traffic signal control by collecting abundant real-time traffic data\cite{Zhu2019BigDA}. The solutions utilizing reinforcement learning (RL) algorithms for online optimization of traffic signal control are proposed in \cite{Wu2017FlowAM,chachatoward,Laval2019LargescaleTS,Guo2019ARL}. It is shown that a potential reduction of up to $73\%$ in vehicle delays can be achieved when compared to fixed-time actuation \cite{Mousavi2017TrafficLC}. 

When the traffic pattern changes dramatically, the state-of-the-art RL algorithms may be unable to make full use of the real-time traffic condition data and give an ineffective adjustment scheme, resulting in longer waiting time of vehicles. Recently, the concept of "max press" (MP) from the traffic field has been utilized as the reward for control model optimization \cite{wei2019presslight, chachatoward}. The pressure is defined as the difference between the number of vehicles on incoming lanes and the number of vehicles on outgoing lanes. However, all the studies above ignore the consideration of road carrying capacity. For example, if there has no vacant space on outgoing lane, the waiting vehicles on the corresponding incoming lane will not be able to pass the intersection even in the green light time. The reward used in such cases will be unable to reflect the effectiveness of the selected action. This will result in an unnecessary waste of green light resources and increase the waiting time of all vehicles that fail to pass.

\begin{figure*}[t]
	\centering
	\includegraphics[width=0.8\linewidth]{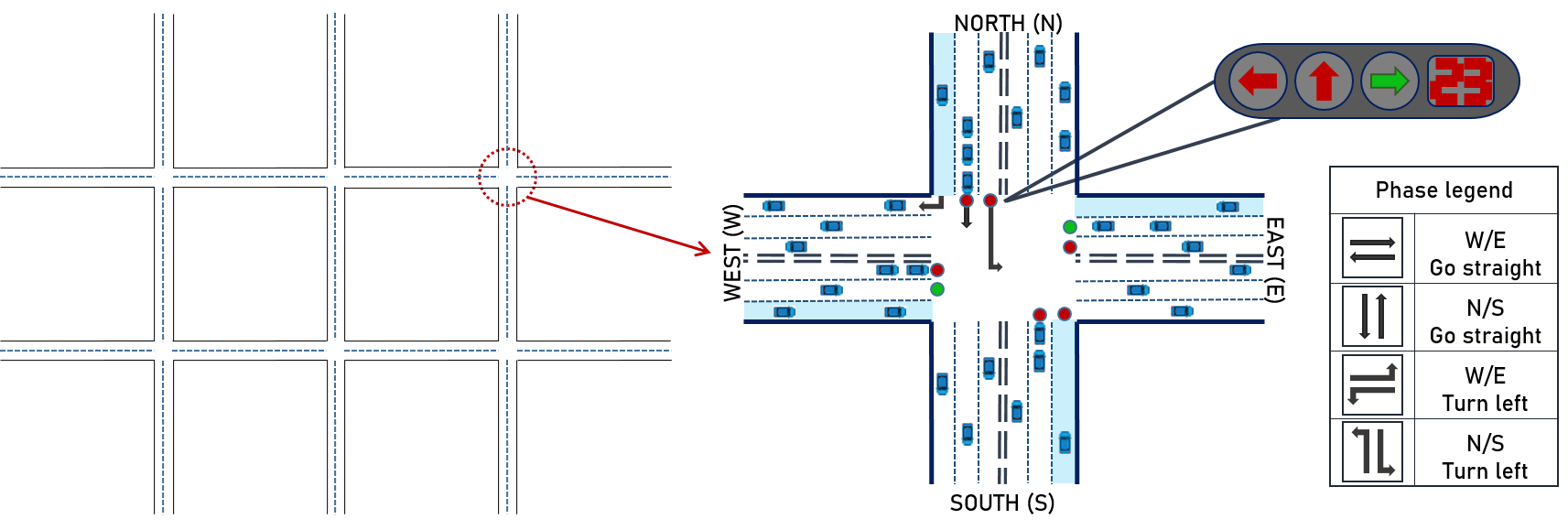}
	\caption{Left: The road network. Right: The traffic lights and phase settings at a single intersection, which have $4$ type of phases and allows right turns at any time.}\label{fig1}
\end{figure*}

To enhance the capacity of urban road network in a dynamic way, our idea is that the computation of the pressure should consider not only the vehicles on the incoming lanes but also the capacity of the outgoing lanes.  %Moreover, the dynamic green light technique sets the duration of a traffic signal according to real-time traffic conditions. 
An RL agent is assigned for each intersection in the traffic network, which observes the real-time traffic state of its own intersection at each time step. Then according to the observed traffic condition, the agent selects an action from the action space and executes it until the next time step. After the execution, state parameters obtained from the traffic conditions such as the number of vehicles on the lane will be given. The performance of the selected action will be judged by the reward and the policy will be optimized.  To summarize, the main contributions are:

\begin{itemize}
\item We propose PRCOL, a more rigorous way of calculating pressure to capture the real-time feature of the incoming and outgoing lanes condition. This approach avoids the phenomenon that vehicles cannot completely pass the intersection due to limited outgoing lane capacity.

\item  We use the  PRCOL as the reward function and design an RL algorithm to control the traffic light in the road network of multi-intersections.

\item We perform simulation experiments under two synthetic data-sets and two real-world data-sets, Hangzhou and New-York. We consider two cases when the traffic duration is Fixed or Dynamic. The results demonstrate that the proposed PRCOL can reduce the travel time of vehicles and increase the throughput of the network.
\end{itemize}

\section{Related Work} \label{sec_2}

Since RL algorithm can learn from the interaction with the environment and adapt to the rapidly changing traffic environment, it achieves better performance than the traditional approaches. There has been a large volume of published studies of RL algorithms in the field of traffic signal control.

As fundamental elements in the RL algorithm, the setting of reward and action will have a significant impact on its performance. Variables which are more easily observed, such as queue length and average delay, are often used as reward parameters, such as in \cite{zangmetalight,zheng2019learning,de2006reinforcement,wei2019colight,chu2019multi,wei2018intellilight,genders2020policy,joo2020reinforcement}. However, these heuristic setting may cause high sensitivity performance and increased learning process. \cite{wei2019presslight} and \cite{chachatoward} propose a reward setting approach based on max pressure (MP) inspired by relevant research in the field of transportation\cite{lioris2016adaptive}. The “pressure” is defined as the difference between the number of vehicles on incoming lanes and outgoing lanes. The MP does not take into account the carrying capacity of the lanes, so that excessive traffic flow may cause system failure. In this paper, we design a novel pressure, PRCOL, and use it as the reward for the RL algorithm.

There are usually three action options for traffic light control problems. A simple way is to have the agents choose whether to switch to the next phase in a cycle-based signal plan\cite{xu2020network,de2006reinforcement,wei2018intellilight}, which is not flexible enough to cope with changing traffic conditions. A most widely-used approach is to select  the green phases for next state \cite{wei2019presslight,chachatoward,zangmetalight,zheng2019learning,wei2019colight,kim2020cooperative,chu2019multi,genders2020policy}. However, fixed action duration may cause unnecessary delay when the vehicle's required pass time and green phase duration do not match. In other approaches, the agents can choose the duration of green lights to adapt to the changing traffic flow\cite{liang2019deep,joo2020reinforcement}. In the experiment, a more flexible approach is used, which can adjust the duration of the traffic light according to the real-time traffic conditions.

\section{Problem Definition}\label{3}

This paper studies the dynamic control of traffic light phase at a multi-intersection scene. Every single intersection is set as a two-way six-lane intersection, as shown in Fig.\ref{fig1}. Each intersection has four approaches in different directions ("W", "E", "N", "S"), which are divided into three lanes (turn left, turn right, go straight). The green light directions are combined in pairs in a way that does not cause conflict, and are divided into four phases, as given in Fig. \ref{fig1}.

%fig1

\paragraph{Incoming/outgoing approach and lane.} Incoming and outgoing approaches are defined as the roads on which vehicles approach and leave the intersection. In our setting, each intersection has $8$ approaches. There are three lanes on each approach, representing three different travel directions: turning left, turning right and going straight.

\paragraph{Traffic movement and phase.} A traffic movement is a route that vehicles take through an intersection. The traffic movement from lane $x$ to lane $y$ is expressed as ($x$,$y$). There are $12$ types of moving traffic at an intersection. Two non-conflicting movements is combined as a light phase, and there are $4$ phases in an intersection, as shown in Figure.\ref{fig1}.

\paragraph{PRCOL.} In this paper, the pressure algorithm  \cite{wei2019colight} has been improved, and the impact of the maximum carrying capacity of the lane on the pressure calculation is taken into consideration. Our proposed pressure index is named as PRCOL(Pressure with Remaining Capacity of Outgoing Lane). The PRCOL of a traffic movement is determined by the number of vehicles on the incoming and outgoing lanes, which is calculated by the following equation:
\begin{equation}
\label{equ_PRCOL}
P_i = N_{in}*(1- \frac{N_{out}}{N_{max}})\ ,
\end{equation}
where $P_i$ is the PRCOL of traffic movement $i$, $N_{in}$ and $N_{out}$ is the number of vehicles on the  incoming and outgoing lane respectively. The $N_{max}$ is the maximum number of vehicles that can fit in a lane. There is a one-to-one correspondence between pressures and traffic movements.

\section{PDlight Algorithm} \label{sec_method}\label{4}

In the RL algorithm, the agent making decisions in the process of interacting with the environment is usually regarded as a Markov Decision Process (MDP), represented by $<S, A, p, R, \gamma>$. In this MDP, there is a set of states $s_t\in S$, a set of actions $a_t\in A$, a transfer probability $p$, a reward function $R$, and a discount factor $\gamma$.

In PDlight algorithm, an agent is set for each intersection in the road network. Each agent controls the traffic light of its intersection. For the communication and coordination between the agents, we adopt the graph attention networks proposed by \cite{wei2019colight}.  

Fo each agent, the state reflects the observation of the traffic  situation at a certain time of the intersection. To be more specific, the state is a $12$ dimension vector representing the number of vehicles on the 12 incoming lanes. For the agent, to choose an action is to control the traffic light for the intersection. There are $4$ possible traffic light phases, as shown in Fig. \ref{fig1}. The phase with the minimum intersection pressure calculated by the PRCOL is selected as the green light of the next time step. Considering that the duration of the traffic light may change, we also include this duration into the action space. The choice of reward is based on the PRCOL, which is defined as:
\begin{equation}
r_i=-P_i\ ,
\end{equation}
where $P_i$ is the PRCOL of the traffic movement $i$, which is defined in Eq.\ref{equ_PRCOL}. The total reward $R$ for an action is:
\begin{equation}
R=\sum{r_i}\ ,
\end{equation}
where $i$ is the index of traffic movements.

Unlike the primary pressure algorithms, the PRCOL takes into account the difference between the apparent pressure and the actual capacity of the vehicle. As a more accurate way of reward calculating, it can avoid conflicts caused by the number of vehicles trying to pass exceeding the maximum carrying capacity of the outgoing lane.

At each intersection, the agent observes the road environment at time $t$, obtains the vehicle queue length on each lane and the current traffic light phase, and takes these obtained traffic flow characteristics as the current state. Then, the agent predicts the Q-value of all phases that can be chosen. The Q function is used to predict the expected future reward for a given state and action:
\begin{equation}
	Q(s_t, a_t)=R(s_t, a_t)+\gamma*\max\{Q(s_{t+1}, a_{t+1})\}\ ,
\end{equation}
where $\gamma$ is the discount factor, evaluated in the interval between $0$ and $1$, representing the importance of the future state. As $\gamma$ approaches $1$, the agent will focus on the reward for the future state.

The goal of action selection is to maximize the reward $R$, that is, to choose the phase in time step $t$ with the highest Q-value. Then, the agent will calculate the maximum number of vehicles that can move across the intersection. The loss function expressed as follow is calculated and the parameters of the Q network are updated by gradient descent.
\begin{equation} \label{equ_J}
	J = \sum \frac{1}{B}\left(R_t+\gamma \max_{a_{t+1}} \hat{Q}\left(s_{t+1},a_{t+1};\hat{\theta}\right)-Q\left(s_t,a_t;\theta\right)\right)^2\ ,
\end{equation}
where $\hat{Q}$ is the target function, and $Q$ is the primary function.

PDlight algorithm uses Deep Q-Network (DQN) to control the traffic lights at each intersection. As one of the classical reinforcement learning algorithms, DQN combines Q-learning with deep neural network, which is used to approximate the Q-value function to avoid the “curse of dimensionality”. In the complex and unknown environment, DQN interacts with the environment by selecting actions and obtaining rewards in a trial-and-error manner, so as to perceive the invisible environment and learn the optimal strategy. The pseudocode of PDlight algorithm is shown in Algorithm \ref{alg:DQN}.

\begin{algorithm}[t]
	\caption{PDlight: Traffic Light Control with Pressure and Dynamic Green Light Time}
	\label{alg:DQN}
	\begin{algorithmic}
		%\Require
		\State \textbf{Input}: replay memory $m$, sample size $B$, episode length $T$, discount factor $\gamma$, greedy $\epsilon$, learning rate $\alpha$, replacement frequency $C$
		\State Initialize $Q$ with parameters $\theta$, $\hat{Q}$ with parameters $\hat{\theta}$
		\For{each $episode$} 
		    \State Initialize step number $t$ as $0$, total time $t_{sum}$ as $0$
		    \While{$t_{sum} < T$}
		    \State Select a random phase  $pha$ with probability $\epsilon$
		    \State Otherwise $pha \gets \mathop{\arg\max}_{pha}{Q(s_t,pha;\theta)}$
		    \State Receive the  green phase duration time $t_{green}$ from the environment
		    \State Execute $a_t \gets \{pha, t_{green}\} $
		    \State Observe the new state $s_{t+1}$
		    \State Calculate PRCOL and the reward $R_t$
		    \State Store transition $(s_t, a_t, R_t, s_{t+1})$ in $m$
		    \State $t_{sum} \gets t_{sum}+t_{green},\ t \gets t+1$
		    \If{$|m| > B$}
		        \State Select $B$ samples from $m$ randomly
		    \EndIf
		    \State Calculate the loss $J$ by Eq. \ref{equ_J} and update $\theta$ by Gradient Descent with learning rate as  $\alpha$
		    \State Every $C$ steps update $\hat{Q} \gets Q$
		    \EndWhile
		\EndFor
	\end{algorithmic}
\end{algorithm}

\section{Experiment and Analysis}\label{5}

In this section, PDlight algorithm is performed on four data-sets which include both synthetic and real-world traffic data. The experiments are conducted on CityFlow\cite{ZhangCityFlow}, a large-scale traffic simulator.

\begin{figure*}[t]
	\centering
	\subfigure[]{\includegraphics[width=0.25\linewidth]{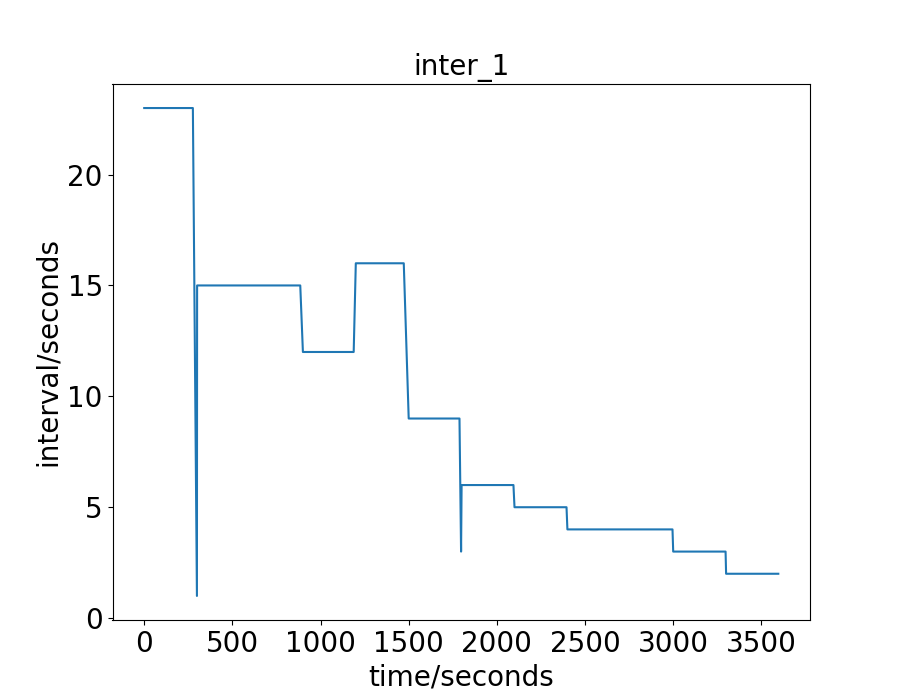}}
	\subfigure[]{\includegraphics[width=0.25\linewidth]{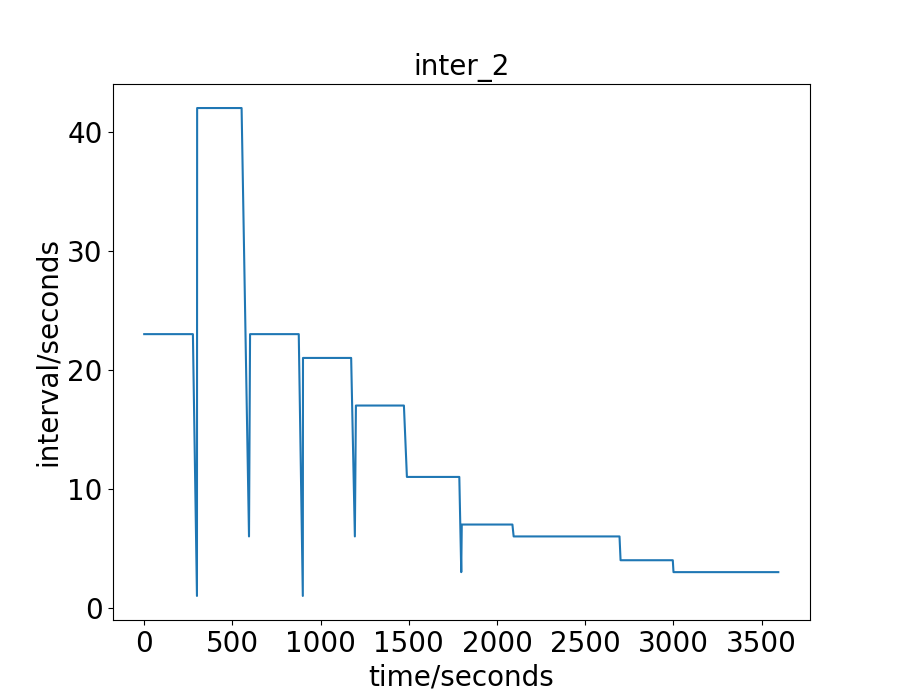}}
	\subfigure[]{\includegraphics[width=0.25\linewidth]{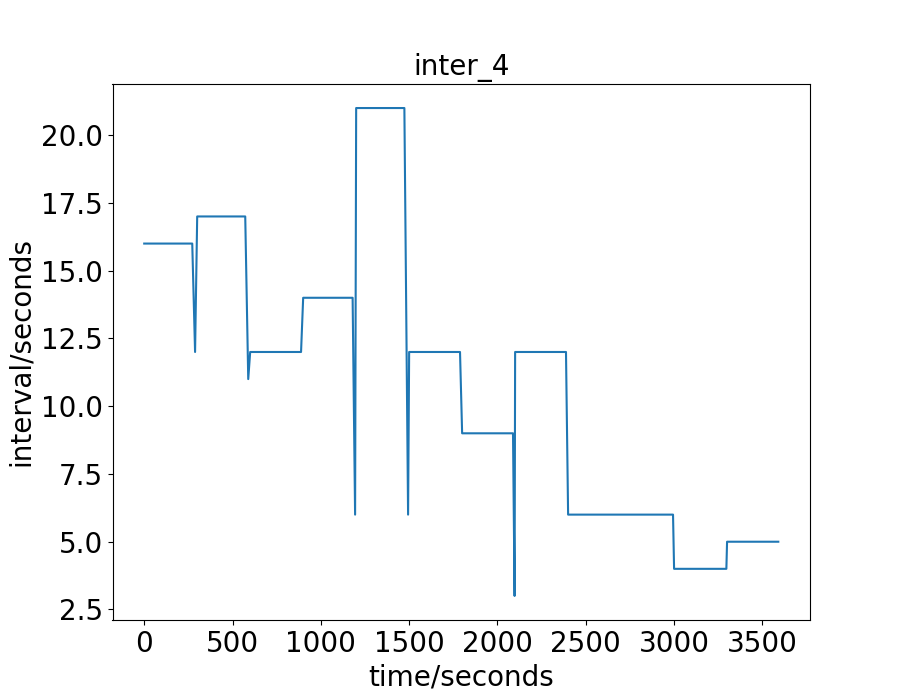}}
	\caption{Arrival interval of the  Hangzhou Data-set }
	\label{fig_hangzhou}
\end{figure*}

\begin{figure*}[t]
	\centering
	\subfigure[]{\includegraphics[width=0.25\linewidth]{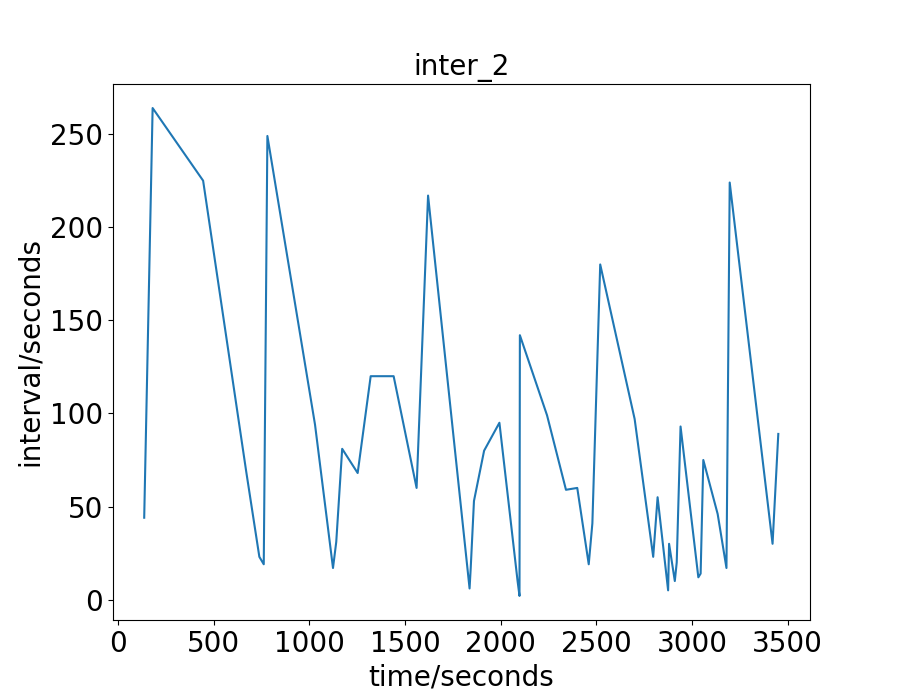}}
	\subfigure[]{\includegraphics[width=0.25\linewidth]{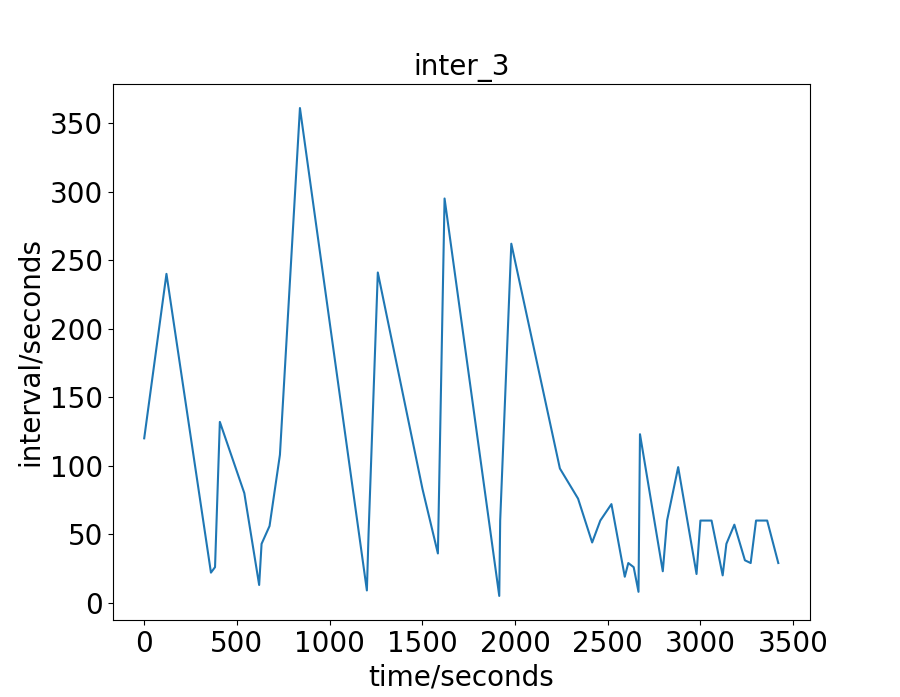}}
	\subfigure[]{\includegraphics[width=0.25\linewidth]{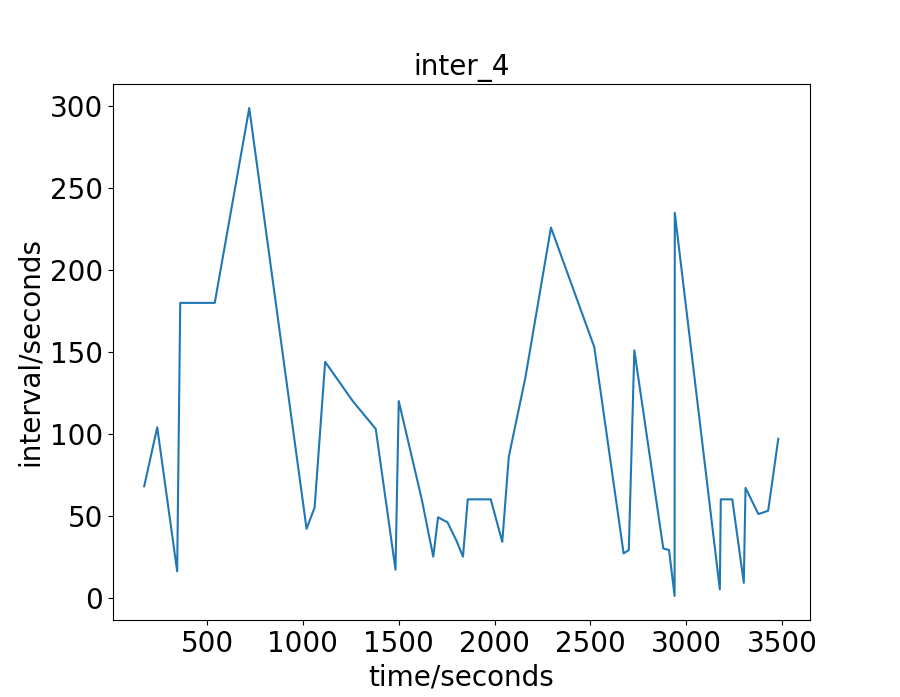}}
	\caption{Arrival interval of the  New-York Data-set }
	\label{fig_newyork}
\end{figure*}

\subsection{Data-sets}

We use two kinds of data-sets, synthetic and real-world. The synthetic data-set enables us to learn the property of different algorithms since the traffic can be designed manually and regularly. The real-world data-sets enable us to test the performance of different algorithms on real traffic flow, which can be more irregular and unpredictable. For all the four data-sets, the topology of the road network is similar to the road presented in Fig. \ref{fig1}.

For the synthetic data-set, there are $3 \times 3 = 9$ intersections (The first and the second $3$ represent there are $3$ roads on the WE and NS direction respectively. For the following Hangzhou and New-York data-sets, we use the similar notation.). The trajectories of vehicles are generated manually. The lane length of the WE lane and NS lane is both $300 \ m$. We use two traffic flow, Syn-Light and Syn-Heavy, to test the performance under different traffic conditions. To facilitate the analysis, we assume  that in both the two data-sets all vehicles can only go straight. Therefore, there are four allowed directions for the traffic: NS, SN, WE, EW. For Syn-Light, the interval between two vehicles is 20 seconds for all of the four directions. For the Syn-Heavy data-set, the interval is set as 10 seconds for the first and third 900 seconds. Then at the second 900 seconds, the NS and SN traffic flow interval is set to 2 seconds while the WE and EW traffic interval is still 10s. At the last 900 seconds, the WE and EW traffic flow interval is set to 2 seconds while the NS and SN interval remains 10s. Such setting is expected to model the case of rush time, when the traffic becomes heavy.

The real-world data-sets are the traffic flow of an hour in the city of Hangzhou and New-York. For the Hangzhou data-set, there are $4 \times 4 = 16$ intersections. The lane length for the WE and NS direction is $800 \ m$ and $600 \ m $ respectively. The vehicle trajectory is captured from the surveillance camera. For the New-York data-set, there are $16 \times 3 = 48$ intersections.  The lane length for the WE and NS direction is $350 \ m$ and $100 \ m $ respectively. The vehicle trajectory is open-source \footnote{https://www1.nyc.gov/site/tlc/about/tlc-trip-record-data.page}. For these two real-world data-sets, the traffic is more irregular and complicate than the synthetic data-sets. To have a more intuitive understand of these real-world two data-sets, we draw their vehicles' arrival interval on some intersections, as shown in Fig. \ref{fig_hangzhou} and \ref{fig_newyork}. The x-axis is the time a vehicle enters the road network, and the y-axis is the interval between this vehicle and the next one. For the Hangzhou data-set, the traffic is light in the beginning and becomes heavier later. For the New-York data-set, the traffic presents no distinct characteristics .

%fig_hangzhou

%fig_newyork

\begin{table*}[t]
	\centering
	\begin{tabular}{l|cccc||cccc}
		\toprule
		& \multicolumn{4}{c||}{Average Travel Time} & \multicolumn{4}{c}{Throughput} \\
		\hline
		& Syn-Light & Syn-Heavy  & Hangzhou & New-York & Syn-Light & Syn-Heavy  & Hangzhou & New-York  \\
		\hline
		FixedTime    & 115.36  & 197.75  & 557.68  & 1846.21 
		& 2094  & 5220  & 3410   & 200\\
		\hline
		MaxPressure  & 116.32  & 196.84  & 406.35  & 421.27  
		& 2096  & 6080  & 4394   & 2373  \\
		\hline
		CoLight      & 94.92   & 197.10  & 358.71  & 177.50  
		& 2109  & 6366  & 4360   & 2699\\
		\hline
		PressLight  & 98.40   & 219.82  & 390.42  & 848.51  
		& 2107  & 6009  & 4359   & 1243\\
		\hline
		PDLight      & \textbf{89.24}   & \textbf{190.10}  & \textbf{340.66}  & \textbf{176.80} 
		& \textbf{2112}  & \textbf{6428}  & \textbf{4485}   & \textbf{2710}\\
		\bottomrule 
	\end{tabular}
	\caption{Average Travel Time (/seconds) and Throughput under Fixed green light duration}\label{table_fixed}
\end{table*}

\begin{table*}[t]
	\centering
	\begin{tabular}{l|cccc||cccc}
		\toprule
		& \multicolumn{4}{c||}{Average Travel Time} & \multicolumn{4}{c}{Throughput} \\
		\hline
		& Syn-Light & Syn-Heavy  & Hangzhou & New-York & Syn-Light & Syn-Heavy  & Hangzhou & New-York  \\
		\hline
		FixedTime    & 115.36 & 197.75  & 557.68  & 1846.21
		& 2094  & 5220  & 3410  & 200\\
		\hline
		MaxPressure  & 116.32 & 196.84  & 406.35  & 421.27
		& 2096  & 6080  & 4394  & 2373  \\
		\hline
		CoLight      & 95.13  & 195.31  & 417.77  & 177.95 
		& 2109  & 6386  & 4183  & 2712\\
		\hline
		PressLight & 98.52  & 208.46  & 380.69  & 897.26
		& 2107  & 6402  & 4348  & 1206\\
		\hline
		PDLight       & \textbf{89.69}  & \textbf{194.25}  & \textbf{352.53}  & \textbf{174.51}
		& \textbf{2112}  & \textbf{6446}  & \textbf{4584}  & \textbf{2721}\\
		\bottomrule 
	\end{tabular}
	\caption{Average Travel Time (/seconds) and Throughput under Dynamic green light duration}\label{table_dynamic}
\end{table*}

\begin{figure*}[t]
	\centering
	\subfigure[Hangzhou - number of vehicles] {\includegraphics[width=0.24\linewidth]{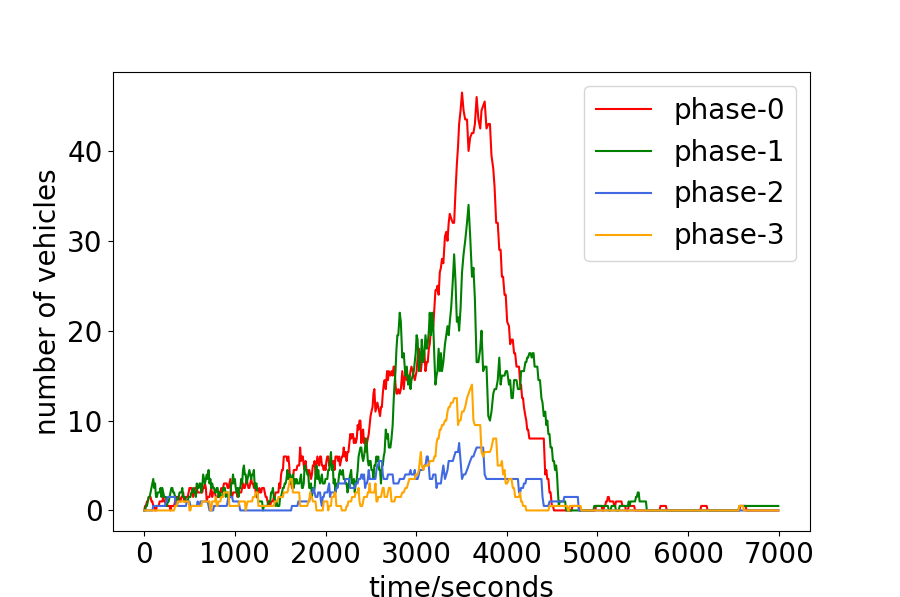}}
	\subfigure[Hangzhou - choice of traffic light] {\includegraphics[width=0.24\linewidth]{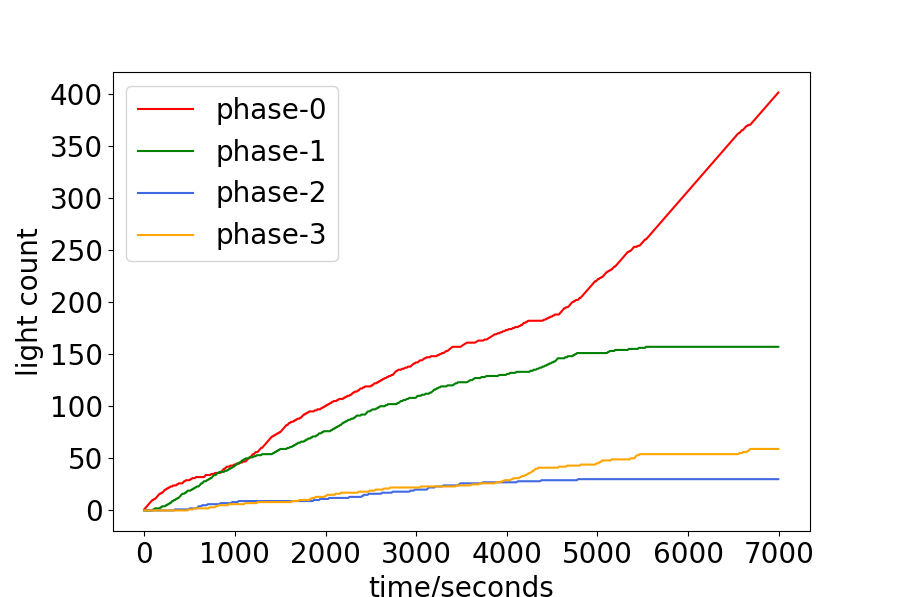}}
	\subfigure[New-York - number of vehicles] {\includegraphics[width=0.24\linewidth]{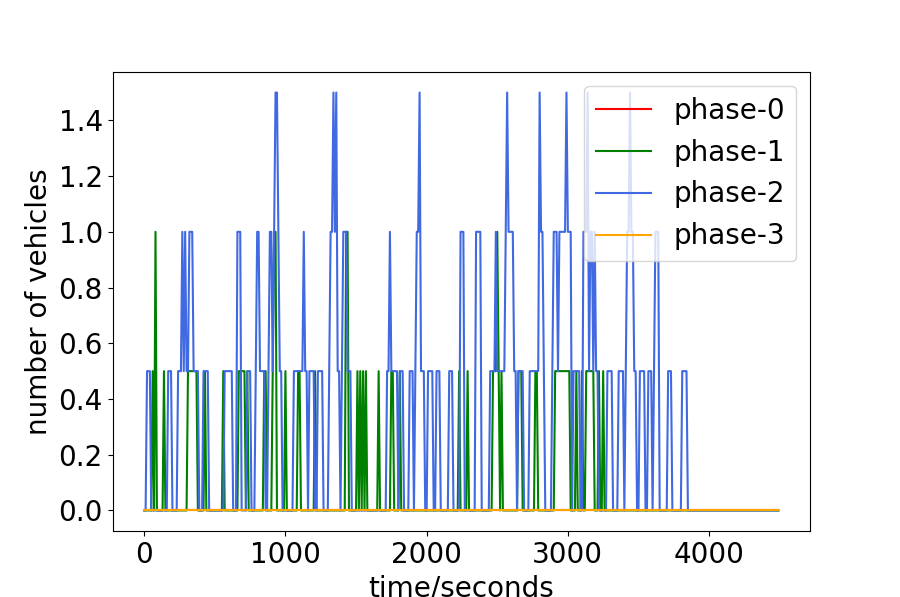}}
	\subfigure[New-York - choice of traffic light] {\includegraphics[width=0.24\linewidth]{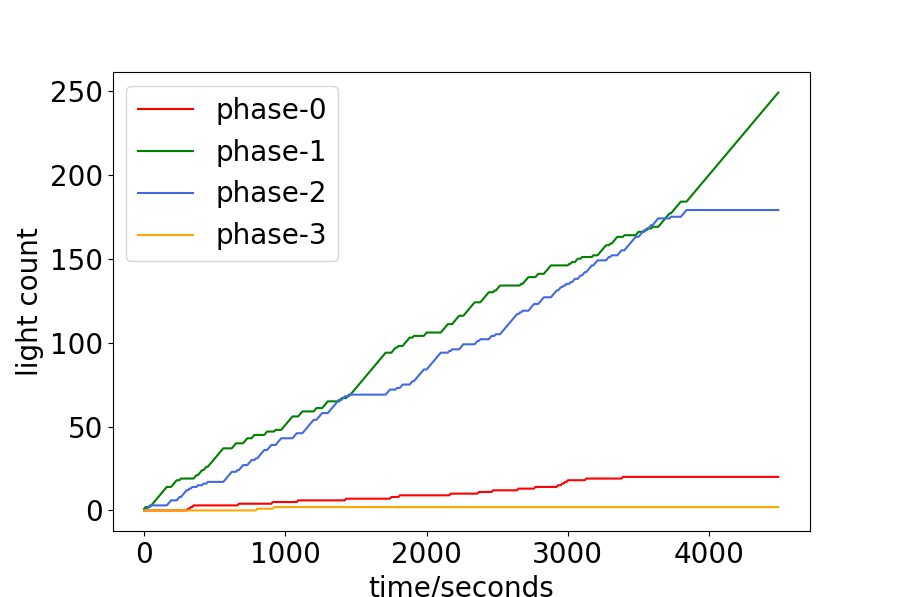}}
	\caption{The number of vehicles and the choice of green light }
	\label{fig_lanenum}
\end{figure*}

\begin{figure*}[t]
	\centering
	\subfigure[2000 - 2500 seconds] {\includegraphics[width=0.4\linewidth]{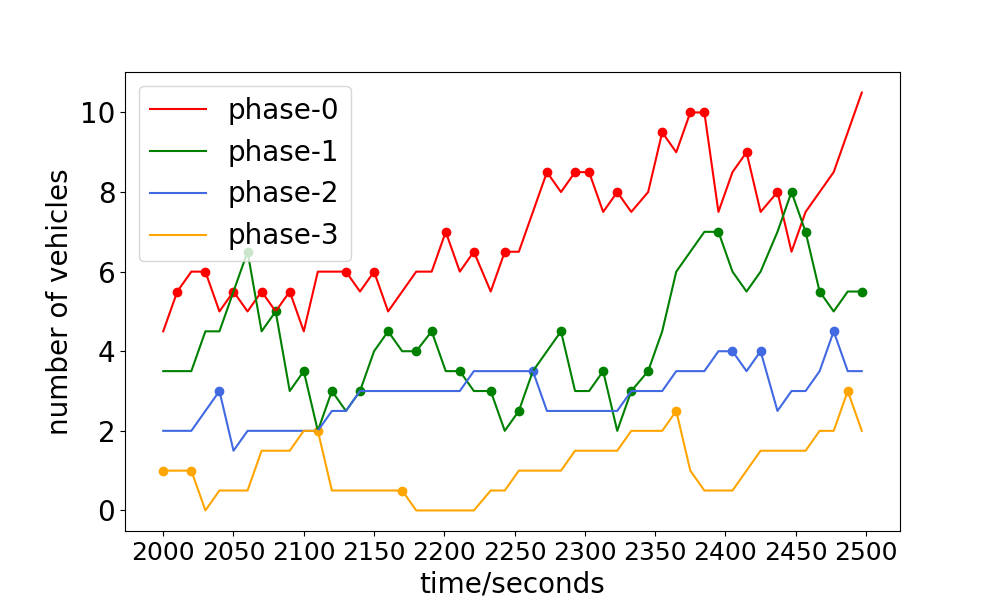}}
	\subfigure[3000 - 3500 seconds] {\includegraphics[width=0.4\linewidth]{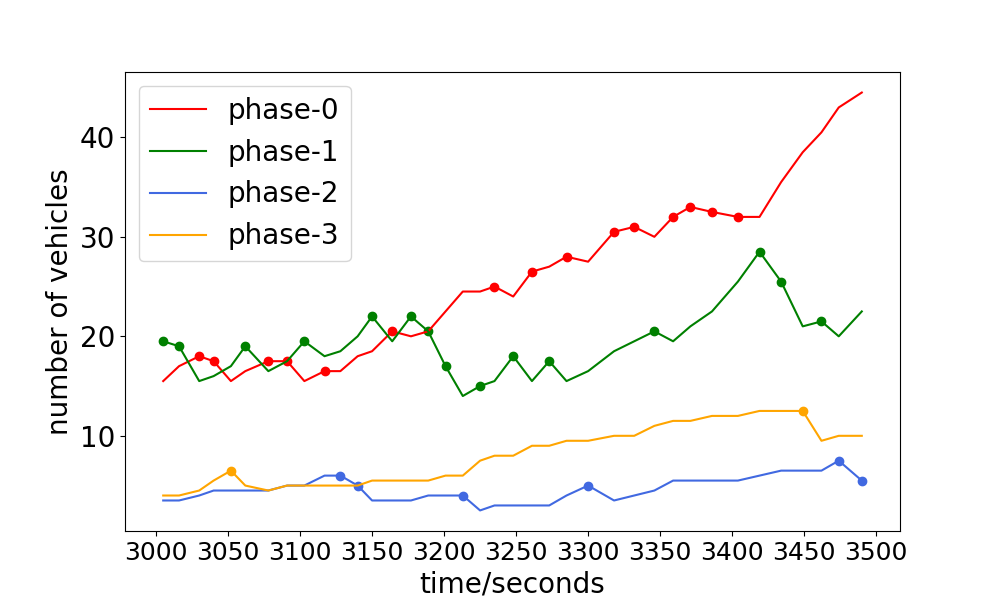}}
	\caption{ The choice of phase in detail on the Hangzhou data-set}
	\label{fig_detail-hangzhou}
\end{figure*}

\begin{figure*}[t]
	\centering
	\subfigure[Syn-Light]{\includegraphics[width=0.24\linewidth]{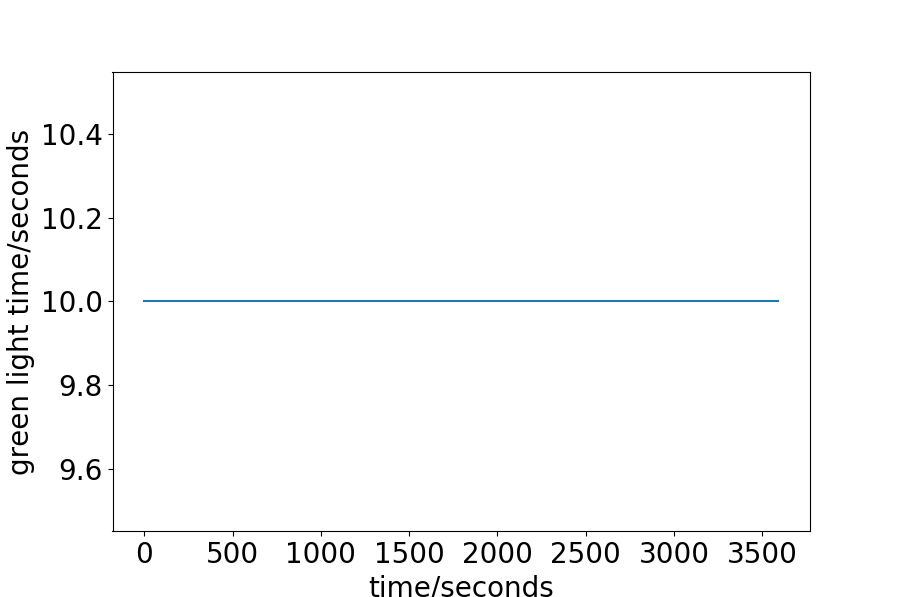}}
	\subfigure[Syn-Heavy]{\includegraphics[width=0.24\linewidth]{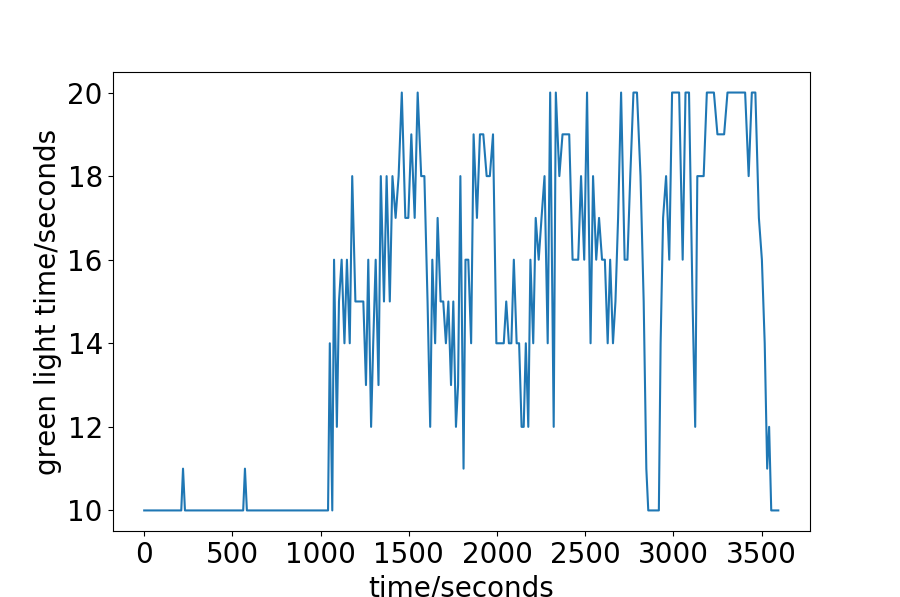}}
	\subfigure[Hangzhou]{\includegraphics[width=0.24\linewidth]{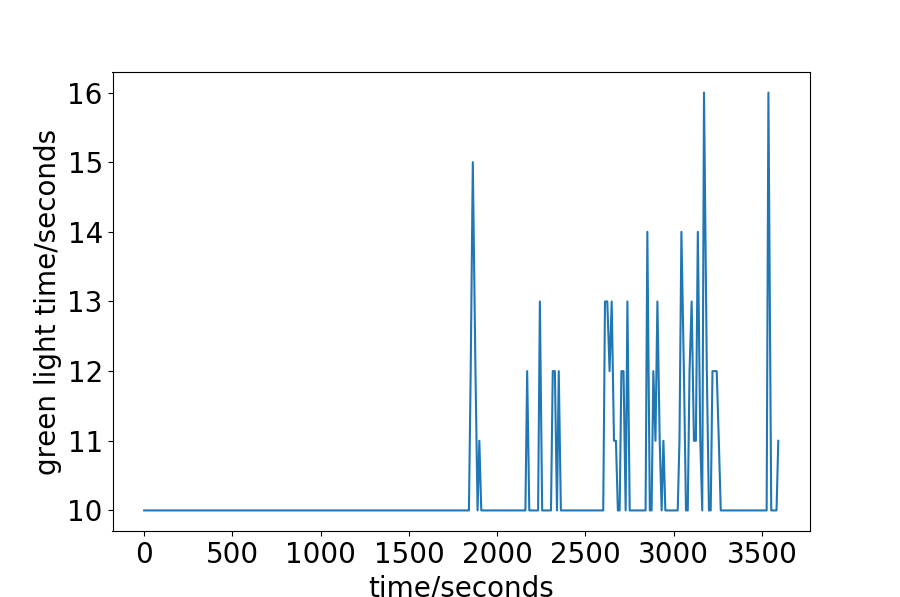}}
	\subfigure[New-York]{\includegraphics[width=0.24\linewidth]{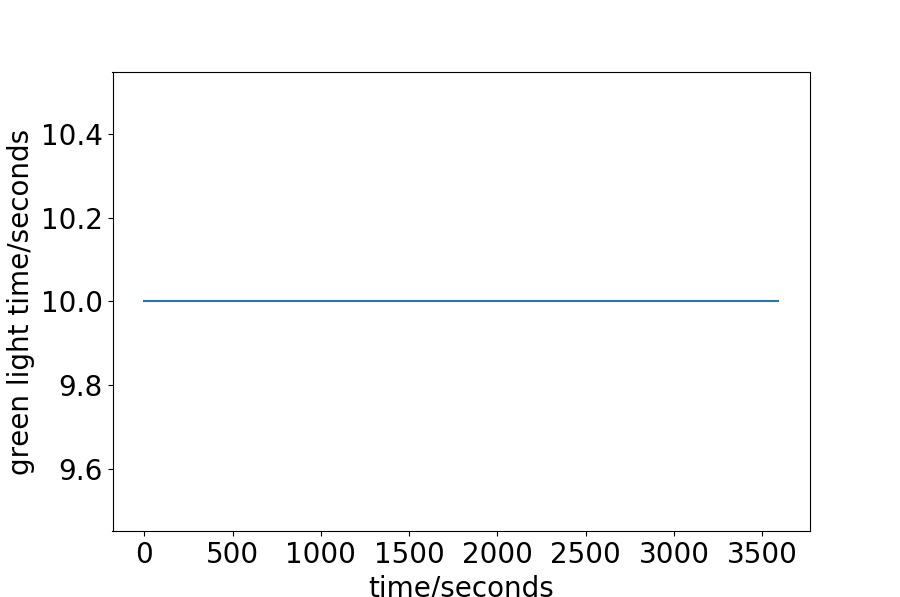}}
	\caption{ The green light time during the process }
	\label{fig_green}
\end{figure*}

\begin{figure*}[t]
	\centering
	\subfigure[inter 1]{\includegraphics[width=0.25\linewidth]{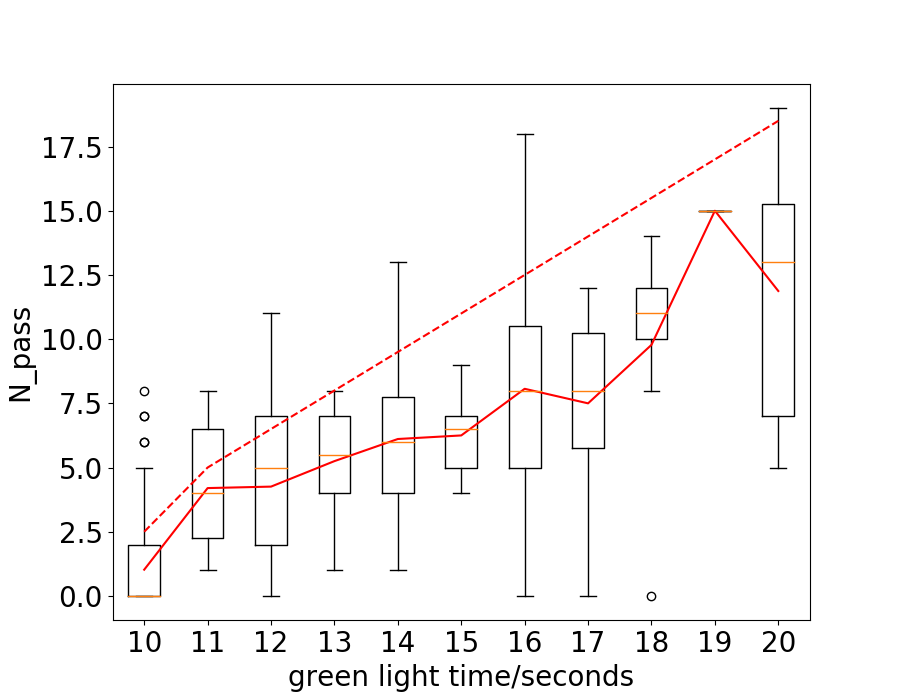}}
	\subfigure[inter 7]{\includegraphics[width=0.25\linewidth]{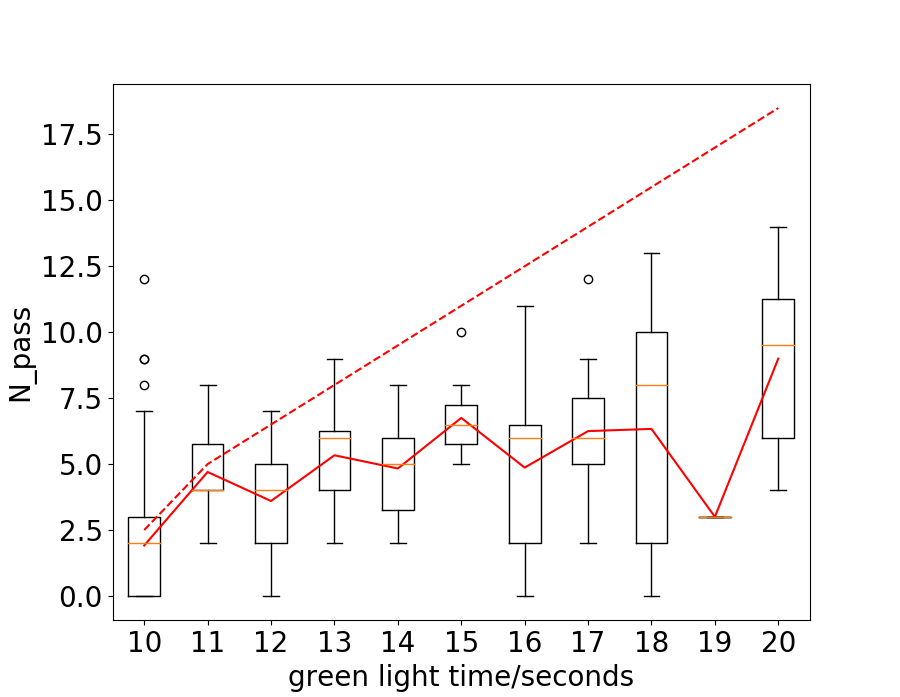}}
	\subfigure[inter 15]{\includegraphics[width=0.25\linewidth]{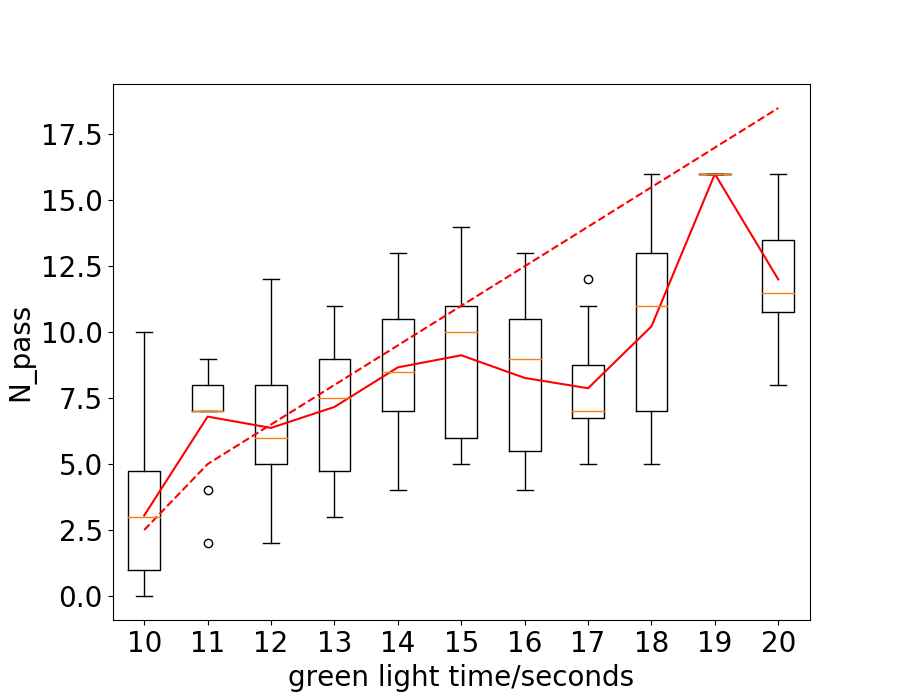}}
	\caption{The number of passed vehicles under different green light time in Hangzhou data-set}
	\label{fig_timeN}
\end{figure*}

\subsection{Experiment Settings}

Regarding the  calculation of the proposed PRCOL in Eq. (\ref{equ_PRCOL}), the number of vehicles on the incoming lane $N_{in}$ and the number of vehicles on the outgoing lane $N_{out}$ can be  measured by some sensors or camera in practice. There are also APIs in CityFlow that can give these measurements. For $N_{max}$, the maximum number of vehicles that can fit in the outgoing lane, we assume that the length of the outgoing lane is $l_l$, the average length of the vehicle is  $l_v$ and the minimum gap between two vehicles  is $l_g$. The $N_{max}$ then can be simply calculate as $\lfloor l_l/(l_v + l_g) \rfloor $. In the experiment, for all the four data-sets, we take $l_v = 5 \  m$ and $l_g = 2.5 \ m$.  

We consider two settings of  traffic light duration: Fixed and Dynamic. For the Fixed traffic light duration, the duration is set as 10 seconds for all green lights. For the Dynamic traffic light duration, we adjust the duration of green light according to the real-time traffic condition. To be more specific, we first define the number of vehicles to pass as 
\begin{equation} \label{equ_Npass}
N_{pass} = \min \{N_{in}, N_{left}\},
\end{equation}
where $N_{left} = N_{max} - N_{out}$ is the number of remaining empty space of the outgoing lane. We assume there are $N_{pass}$ vehicles waiting before the intersection statically and their gap is the minimum gap $l_g$. With the acceleration as $a$ and the maximum speed as $v$, we can calculate the total time for all the $N_{pass}$ vehicles to pass the intersection. In the experiment, we take  $a = 2 \ m/s^2$ and  $v = 40 \ km/h$. For both the Fixed and Dynamic traffic light duration, a yellow light of $5$ seconds is added to clear the traffic and avoid collision whenever the green light swift to another phase. 

In algorithm \ref{alg:DQN}, the episode length is set as $3,600$ seconds. In the Fixed traffic light duration, each $t_{green}$ is $10$ seconds. And in the Dynamic traffic light duration, the $t_{green}$ ranges from $10$ to $20$ seconds. Note that the actual episode may last longer than one hour in the Dynamic setting. In the experiment, we take $\epsilon$ decreasing from $0.8$ to $0.2$. The discount factor $\gamma$ for calculating the accumulated reward is set as $0.8$. The maximal sample size is $10,000$. The Q network parameters $\theta$ are updated with the learning rate as $0.001$. The target Q network $\hat{\theta}$ is updated every $5$ steps.

%and batch size $20$

\subsection{Baseline}
To demonstrate the effectiveness of the proposed PRCOL, we use both classic traffic light control algorithms and recently developed RL algorithms as baselines.

For the classic traffic light control algorithms, we consider the following two widely used ones:
\begin{itemize}
	\item \textbf{FixedTime}: Set green light for all phases  periodically with a given order.
	\item \textbf{MaxPressure} \cite{Varaiya2013MaxPressure}: Set the green light for the phase with the maximum pressure.
\end{itemize}

Recently, some RL algorithms have been proposed for more advanced traffic light control strategies. In this paper, we use one state-of-the-art framework CoLight proposed in \cite{wei2019colight}. We compare the following three RL algorithms:

\begin{itemize}
	\item \textbf{CoLight} \cite{wei2019colight}. A multi-intersection traffic control algorithm, which uses graph attention networks for the communication among intersections. This algorithm can handle the traffic signal control in a road network containing many intersections. Follow the same setting of  \cite{wei2019colight}, the queue length of the incoming lane  is used as the reward. 
	\item \textbf{PressLight} \cite{wei2019presslight}. A recently developed RL algorithm with pressure as the reward function. In this paper, we use the CoLight framework with pressure as the reward function to accentuate the effectiveness of the proposed PRCOL.
	\item \textbf{PDLight}. The CoLight framework is adopted for the agent and the proposed PRCOL is used to calculate the reward.
\end{itemize}

For the RL algorithms, considering the randomness of the initial value of neural networks, we run each experiment three times with random initial values.

\subsection{Evaluation Metrics} To evaluate the performance of a traffic light control strategy, we use both the travel time and throughput as metrics.

\begin{itemize}
	\item \textbf{Average Travel Time}. In the experiment, we record $t_e$ and $t_l$, the time when a vehicle enters and leaves the road network. The travel time for that vehicle is then $t_l - t_e$. Since not all vehicles can leave the network in the given one hour, for those left in the network, we calculate their travel time as  $T - t_e$. We use the average travel time for all vehicles as the final result.
	\item \textbf{Throughput}. The throughput is defined as the number of vehicles passed the road network in the give one hour. Each vehicle has a pre-determined destination and only the vehicles that have reached the destination are counted as passed.
\end{itemize}

\subsection{Result}

In Table \ref{table_fixed}, we give the average travel time for all vehicles  and the throughput under Fixed green light duration.  In Table \ref{table_dynamic}, we give the average travel time for all vehicles  and the throughput under Dynamic green light duration. It can be seen that under both traffic light settings, the proposed PDlight achieves the lowest average travel time and the highest throughput.

%table_fixed

%table_dynamic

\subsection{Case Study}
The average travel time and throughput has demonstrated the superior of the proposed PDLight. To gain a more comprehensive understanding of the algorithm, we will analyze the action chosn by the PDlight algorithm in detail.

The PDlight aims to control the traffic light in the complex and dynamic traffic environment according to the real-time traffic condition, and thus we analyze the relationship between the traffic condition and the choice of traffic light. The number of vehicles of the four phases and the accumulated number each phase has been chosen are shown in Fig. \ref{fig_lanenum}. Since each phase corresponds to two directions, the number of vehicles is calculated by their average. (For example, phase-0 means the traffic flow goes straight in W/E will be allowed, as show in  \ref{fig1}, and the number of vehicles of phase-0 is the average of the number in the west-to-east lane and the east-to-west lane.) From \ref{fig_lanenum}.(a) and \ref{fig_lanenum}.(b), or \ref{fig_lanenum}.(c) and \ref{fig_lanenum}.(d), it can be seen that the more vehicles of one phase, the more time it will be chosen, such as phase-0 of the Hangzhou data-set and phase-1 and phase-2 of the New-York data-set. Quantitatively, we calculate the frequency that the phase with the maximum real-time number of vehicles has been chosen in Hangzhou and New-York data-sets as \textbf{0.518} and \textbf{0.549} respectively.

%fig_lanenum

The experiments have shown that with more than $50\% $, the phase with the maximum number of vehicles is chosen, but \textit{When will the algorithm choose the phase not with the maxinum number of vehicles?} To answer this question, the number of vehicles and the choice of traffic light are drawn in Fig. \ref{fig_detail-hangzhou}. The point on each curve means one choice of the corresponding phase. It can be seen that for the phase-2 or phase-3, the interval between two choices is relatively long, compared to phase-0 and phase-1. One rational explanation for this can be stated as follows. The aim of traffic light control is to minimize the \textit{average} travel time of all vehicles. For a same interval, green light on the phase with more number of vehicles means less average results. Therefore, the phase with maximum number of vehicles will be chosen priorly. However, if the vehicles of one phase have been waiting long enough, it will be reasonable to set the green light for this phase since some extremely huge numbers may have a deep impact on the result. 
%fig_detail-hangzhou

\subsection{Verification of the dynamic green light duration}

We mention earlier that we consider two traffic light duration: Fixed and Dynamic. The Fixed green light is commonly seen in the daily life. The Dynamic green light duration is also by some previous work. In this subsection, we investigate the future of the duration of green lights. 

We first give the exact green light time during the process in Fig. \ref{fig_green}. For the Synthetic-Light data-set, the interval between two vehicles is set to 20 seconds, and the number of vehicles to pass is relatively small, so the green light is always 10 seconds. For the Hangzhou data-set, as has been discussed, there is a trend for decreasing intervals, or increasing vehicles. The green light time, given this property, also shows some increasing trend. For the approximate first half an hour, the green light is always set to 10 seconds, which is in consist with the analysis of the data-set aforementioned. For the second half an hour, since the interval becomes smaller, more vehicle enters the road network, and the green light time is set longer than before. This explanation is also adaptive to the Synthetic Heavy data-set. For the New-York data-set, the green light is also set as 10 seconds constantly. The analysis in Fig.\ref{fig_newyork} shows no clear trend as the Hanzhou or Synthetic data-sets. The range of the intervals  is wide, from less than 5 to more than 350 seconds. One can also notice that the small interval is less frequent than the large interval. The average interval is calculated to be 42.55 seconds, which is considerably large.
%fig_green

The duration of the green light is decided by $N_{pass}$, the number of vehicles we assume will pass the intersection. It is natural to ask what is the real number of vehicles passing the intersection with the given green light time. To answer this question, we take Hangzhou data-set as an example and draw the curve of passed vehicles vs green light time in Fig. \ref{fig_timeN}. The solid and dotted red lines represent the number of vehicles passed the intersection ideally and practically. It can be seen that the solid line goes as roughly approximate trend as the dotted line,  while always lower than the dotted line. The similar trend validates our assumption and the decision of green light time. The gap between the two lines may be caused by the complex traffic condition in practice. It is noticeably that the vehicles on one lane may change to another while driving and the distance between two vehicles is hardly equal to the minimal threshold for safety. This can also help explain why the gap between the solid and dotted line increases as the duration of green light increases, since longer green light means more uncertainty and  bigger difference between ideal and practice.

%fig_timeN
 
\section{Conclusion}
\label{6}
In this paper, we propose PDlight, a traffic light control algorithm with a novel PRCOL as the reward. The new pressure PRCOL considers not only the vehicles on the incoming lanes but also the capacity of the outgoing lanes. Experiment results on synthetic and real-world data-sets  show that the average travel time of all vehicles in the road network is shorten and that the total trough output is larger. A deep analysis of the  decision of traffic signal  reveals what the PDlight  has learned and can help to design traffic light control algorithms.

\section*{Broader Impact}
Traffic light control is an important technology in intelligent traffic system. The research results in this paper can effectively reduce vehicles' travel time and increase network's throughput, so as to reduce congestion and environmental pollution.

%\small
\bibliography{mycite}

\end{document}